\documentclass[table]{article}

\usepackage{arxiv}

\usepackage{amsmath}

\usepackage[utf8]{inputenc} 
\usepackage[T1]{fontenc}    
\usepackage{hyperref}       
\usepackage{url}            
\usepackage{booktabs}       
\usepackage{amsfonts}       
\usepackage{nicefrac}       
\usepackage{microtype}      
\usepackage{cleveref}       
\usepackage{graphicx}
\usepackage[numbers, square]{natbib}
\usepackage{doi}

\usepackage{tikz}
\usetikzlibrary{fadings}
\usepackage{tcolorbox}
\usepackage{mathtools}
\usepackage{amsthm}
\usepackage{bbm}
\usepackage{bm}
\usepackage{tabularx}
\usepackage{threeparttable}
\usepackage{makecell}
\usepackage{multirow}
\usepackage{booktabs}
\usepackage{subcaption}
\usepackage{flushend}
\usepackage{dirtytalk}
\usepackage{xcolor}
\usepackage{circuitikz}
\usepackage{amsmath}
\usepackage{algorithm}
\usepackage{algorithmic}
\usepackage{enumitem}
\usepackage{hyperref}

\title{A Survey on Current Trends and Recent Advances in Text Anonymization}

\date{}

\usepackage{authblk}

\author[1,2]{%
	Tobias Deu{\ss}er\thanks{\texttt{tdeusser@uni-bonn.de}, ORCID-ID: 0000-0003-4685-0847}%
}
\author[1]{%
	Lorenz Sparrenberg
}
\author[1,2]{%
	Armin Berger
}
\author[1,2]{%
	Max Hahnb{\"u}ck
}
\author[1,2]{%
	Christian Bauckhage
}
\author[1,2]{%
	Rafet Sifa
}
\affil[1]{University of Bonn, Bonn, Germany}
\affil[2]{Fraunhofer IAIS, Sankt Augustin, Germany}

\renewcommand{\headeright}{}
\renewcommand{\undertitle}{An arxiv Preprint}
\renewcommand{\shorttitle}{A Survey on Current Trends and Recent Advances in Text Anonymization}

\hypersetup{
pdftitle={A Survey on Current Trends and Recent Advances in Text Anonymization},
pdfsubject={cs.CL, cs.AI},
pdfauthor={Tobias Deu{\ss}er, Lars Hillebrand, Armin Berger, Christian Bauckhage, Rafet Sifa},
pdfkeywords={Anonymization, Large Language Models, Named Entity Recognition, Natural Language Processing, Privacy, Trustworthy Machine Learning, Survey},
}

\newcommand\copyrighttext{%
  \footnotesize \textcopyright 2025 IEEE. Personal use of this material is permitted. Permission from IEEE must be obtained for all other uses, in any current or future media, including reprinting/republishing this material for advertising or promotional purposes, creating new collective works, for resale or redistribution to servers or lists, or reuse of any copyrighted component of this work in other works.

}
\newcommand\copyrightnotice{%
\begin{tikzpicture}[remember picture,overlay]
\node[anchor=south,yshift=10pt] at (current page.south) {\fbox{\parbox{\dimexpr\textwidth-\fboxsep-\fboxrule\relax}{\copyrighttext}}};
\end{tikzpicture}%
}

\begin{document}
\maketitle
\addtocounter{footnote}{-1}
\begin{abstract}
The proliferation of textual data containing sensitive personal information across various domains requires robust anonymization techniques to protect privacy and comply with regulations, while preserving data usability for diverse and crucial downstream tasks. This survey provides a comprehensive overview of current trends and recent advances in text anonymization techniques. We begin by discussing foundational approaches, primarily centered on Named Entity Recognition, before examining the transformative impact of Large Language Models, detailing their dual role as sophisticated anonymizers and potent de-anonymization threats. The survey further explores domain-specific challenges and tailored solutions in critical sectors such as healthcare, law, finance, and education. We investigate advanced methodologies incorporating formal privacy models and risk-aware frameworks, and address the specialized subfield of authorship anonymization. Additionally, we review evaluation frameworks, comprehensive metrics, benchmarks, and practical toolkits for real-world deployment of anonymization solutions. This review consolidates current knowledge, identifies emerging trends and persistent challenges, including the evolving privacy-utility trade-off, the need to address quasi-identifiers, and the implications of LLM capabilities, and aims to guide future research directions for both academics and practitioners in this field.
\end{abstract}
\copyrightnotice


\section{Introduction}
The digital age has led to an unprecedented generation and collection of textual data \cite{rivery2025bigdata}, from electronic health records and legal documents to social media posts and customer reviews. While this data holds immense value for research, analytics, and service improvement, it often contains sensitive personal information, posing significant privacy risks. Regulatory frameworks like GDPR \cite{Oksanen2022anoppi, CabreraDiego2024psilence} mandate the protection of such data, making effective anonymization techniques indispensable.

Text anonymization aims to transform textual data in such a way that individuals cannot be re-identified, either directly or indirectly, while minimizing the loss of information utility for downstream tasks like contradiction detection \cite{deusser2023uncovering}, algorithmic trading \cite{yu-etal-2023-harnessing}, speech-based dementia detection \cite{deusser2024fusing}, regulatory compliance verification \cite{berger2023towards}, or legal contract analysis \cite{watson-etal-2025-law}. This involves complex challenges, including the accurate detection of diverse personally identifiable information (PII) types, handling contextual ambiguities, preserving semantic integrity, and balancing the often-competing goals of privacy and utility. The advent of sophisticated analytical tools, particularly Large Language Models (LLMs), has further complicated this landscape, offering both powerful new anonymization capabilities and potent de-anonymization threats.

Our contributions are threefold. First, we provide a holistic survey that bridges foundational NER-based anonymization techniques with the emerging role of LLMs, highlighting their dual potential as both anonymization tools and de-anonymization threats. Second, we offer a broad synthesis of domain-specific challenges and solutions across healthcare, law, finance, and education, while also dedicating focused analysis to advanced privacy-preserving methodologies such as formal privacy models and authorship anonymization. Third, we emphasize the importance of robust evaluation, reviewing current benchmarks, metrics, and practical toolkits, and critically assessing the evolving trade-off between privacy and utility. This comprehensive perspective aims to guide both researchers and practitioners navigating the complex landscape of textual data anonymization.

The rest of this paper is structured as follows. We start by discussing foundational approaches, primarily centered around Named Entity Recognition (NER), which have long served as the cornerstone for PII identification. We then delve into the transformative impact of LLMs, examining their dual role as both sophisticated anonymizers and potential re-identification adversaries. Subsequent sections navigate the nuanced landscape of domain-specific anonymization challenges and tailored solutions, particularly in critical sectors like healthcare, finance, and law. We further explore advanced methodologies that incorporate formal privacy models (e.g., Differential Privacy) and risk-aware frameworks. The specialized subfield of authorship anonymization, focused on obscuring linguistic style, is also addressed. Finally, we emphasize the critical importance of robust evaluation frameworks, comprehensive metrics, and the development of practical toolkits and systems for real-world deployment. This review aims to consolidate current knowledge, highlight emerging trends and persistent challenges, and guide future research directions for both academics and practitioners in the field of anonymization.

\section{Foundational and NER-Driven Anonymization}

Named Entity Recognition (NER) has long been a cornerstone of automated text anonymization, serving as the primary mechanism for identifying explicit mentions of PII such as names, locations, organizations, and contact details \cite{deusser2024informed}. Many early and ongoing efforts build upon NER, often augmenting it with rule-based systems, gazetteers, and traditional machine learning techniques.

For specific applications like call centers, \cite{Kaplan2020may} demonstrated the successful application of custom NER models (BiLSTM-CRF with contextual string embeddings) to noisy call-center transcripts for PII masking and privacy law compliance.

In educational contexts, \cite{Bosch2020hello} presented a method using set operations and filtering candidate private words (annotated manually or via ML) to redact PII in online discussion forums, achieving high recall. Similarly, \cite{Farrow2023names} focused on anonymizing participant names in online discussions, including nicknames and spelling errors, using a pseudonymization process guided by class lists, NER, and heuristic rules, arguing that such guided methods can outperform deep neural networks for specific name variant challenges.

The development of practical tools often relies heavily on these foundational techniques. For instance, ANOPPI, a tool for semi-automatic anonymization of Finnish legal texts, combines rule-based, machine-learning-based, and gazetteer-based methods for NER, assigning identifiers for consistent pseudonym replacement \cite{Oksanen2022anoppi}. Textwash, an open-source Python tool, also combines NER with pattern matching and fuzzy string matching to identify and mask PII in a language-agnostic manner \cite{Kleinberg2022textwash}.

These approaches highlight the importance of NER as a fundamental building block, often forming the first pass in more complex anonymization pipelines. However, they also underscore the limitations of relying solely on NER, particularly in handling implicit identifiers, contextual nuances, and ensuring comprehensive privacy protection against sophisticated re-identification attacks, paving the way for more advanced methodologies.

\section{Anonymization for and via LLMs}

The advent of LLMs has significantly impacted the field of text anonymization, offering new capabilities for both identifying and transforming sensitive information, as well as posing new challenges as potential de-anonymization tools.

Several works explore using LLMs directly as anonymizers. \cite{Staab2024largeICLRWorkshop} and \cite{Staab2024largeICLR} investigated whether suitably prompted off-the-shelf LLMs (like GPT-3.5/4) can serve as effective zero-shot or few-shot text anonymizers. Their findings suggest that LLMs can remove or replace identifiable spans while preserving fluency, though consistency and completeness remain challenges. \cite{Deusser2025resource} proposed a novel approach to distill knowledge from LLMs into smaller, more resource-efficient encoder-only models for anonymization, using NER and regular expressions, thereby enabling deployment on less powerful devices.

In the medical domain, LLMs have shown particular promise. \cite{Liu2023deidgpt_arxiv} introduced DeID-GPT, a framework leveraging GPT-4 for zero-shot de-identification of medical texts, reporting high accuracy in masking PHI while preserving text structure. \cite{Altalla2025evaluating} further evaluated GPT-3.5 and GPT-4 for clinical note de-identification, with GPT-4 demonstrating superior performance, highlighting its potential for safeguarding patient privacy.

LLMs are also being used to develop more robust and utility-preserving anonymization frameworks. \cite{Yang2024rupta} proposed RUPTA, a framework using LLMs as a privacy evaluator, utility evaluator, and optimizer to iteratively edit text for an optimal privacy-utility trade-off. \cite{Frikha2024incognitext} introduced IncogniText, an LLM-based method for conditional text anonymization that randomizes private attributes at the document level to prevent inference attacks while preserving meaning. 

However, the power of LLMs also presents a new threat. \cite{Patsakis2023man} highlighted that LLMs like GPT-3.5 can act as effective "de-anonymizers," inferring identities from contextual information remaining after standard anonymization, calling for a re-evaluation of existing techniques. This dual role of LLMs—as both powerful anonymization tools and sophisticated adversaries—is a central theme in current research.

\section{Domain-Specific Challenges and Solutions}

Different domains present unique challenges for text anonymization due to varying data characteristics, types of sensitive information, and regulatory requirements. Researchers have developed tailored solutions for several key areas.

\subsection{Healthcare and Clinical Notes}
The healthcare domain is particularly sensitive, with strict regulations like HIPAA governing Protected Health Information (PHI). De-identifying clinical free text is crucial, and recent approaches have predominantly utilized machine learning. A systematic review by \cite{Kovacevic2024review} covering 69 studies from 2010 to 2023 found that ML and hybrid (machine learning combined with rule-based) methods are the most common, with purely rule-based techniques becoming rare. This review also highlights ongoing challenges such as handling diverse data sources and balancing privacy with data utility.

Transformer-based models have shown significant promise. For instance, \cite{Liu2023transformerdeid} presented Transformer-DeID, which employed models like BERT \cite{devlin-etal-2019-bert} and RoBERTa \cite{liu2019roberta} for de-identifying clinical notes. Comparative evaluations, such as the one by \cite{meaney2022comparativeevaluationtransformermodels}, have explored the performance of various transformer architectures on benchmark datasets, often noting high overall accuracy but with performance variations for specific PHI categories. The generalizability of these models is a critical concern; \cite{chen2024examining} investigated the performance of pretrained de-identification transformers on narrative nursing notes, which can have different characteristics compared to other clinical texts like discharge summaries.

Several established tools and methodologies continue to be relevant. \cite{Ribeiro2023incognitus} introduced INCOGNITUS, a flexible platform for anonymizing clinical notes that offers multiple techniques, including one designed to guarantee 100\% recall by substituting words with semantically similar ones, alongside a module for assessing information loss. \cite{Kocaman2023beyond} detailed a hybrid context-based model for the fully automated de-identification of over a billion clinical notes, which reportedly outperformed both NER-only models and commercial services. Ensemble learning approaches have also demonstrated effectiveness; \cite{Murugadoss2021building} created a "best-in-class" automated de-identification tool for Electronic Health Records (EHRs) by combining rule-based methods, dictionary lookups, and ML models.

The advent of LLMs has opened new avenues. As previously noted, DeID-GPT \cite{Liu2023deidgpt_arxiv} and the research by \cite{Altalla2025evaluating} specifically apply LLMs to the de-identification of clinical text. \cite{Pissarra2024anonymization} conducted a comparative study investigating LLMs for clinical text anonymization, introducing novel evaluation metrics for generative anonymization and concluding that LLM-based methods are a reliable alternative to traditional approaches, achieving high accuracy while preserving utility. To address the poor cross-institutional generalization of de-identification models, \cite{Kim2024privacysafe} proposed using GPT-4 for privacy-safe data augmentation, generating synthetic clinical text (with PHI redacted before prompting) to improve F1 scores on unseen hospital data. Further exploring LLM utility, \cite{Wiest2024.06.11.24308355} introduced the LLM-Anonymizer, a tool using locally deployable LLMs to enhance privacy via on-premise processing. An alternative strategy involves "LLMs-in-the-loop" to develop expert small AI models for de-identification across multiple languages, which may offer advantages in accuracy and privacy over general-purpose LLMs \cite{gunay2024llmsintheloop2expertsmall}.

However, de-identification alone may not suffice for robust privacy. \cite{Sarkar2024synthetic} demonstrated that de-identified clinical notes can still be vulnerable to re-identification (e.g., via membership inference attacks) and explored state-of-the-art LLM-generated synthetic notes as an alternative, finding that while synthetic data can match real data utility, high-fidelity synthetic data can also carry similar privacy risks. Protecting privacy during the training of clinical language models is another critical area. \cite{Vakili2024pseudonymization} applied end-to-end pseudonymization to text data before training a Swedish Clinical BERT, finding minimal to no performance loss on downstream NLP tasks compared to models trained on identified data, suggesting that training on fully de-identified data is feasible. 

Finally, understanding the impact of these techniques is crucial. \cite{Larbi2023clinical} systematically analyzed how various anonymization techniques affect downstream NLP task performance and the risk of re-identification in the medical domain, offering valuable guidance for practitioners.

\subsection{Legal Documents}
Legal documents, such as court decisions and parole hearing transcripts, often contain highly sensitive personal data, necessitating robust anonymization. The systematization of anonymizing court decisions has advanced, particularly with EU Member States' courts adopting algorithmic approaches to automate the process; these efforts also address inherent challenges like re-identification risks and ensuring system acceptance by court staff \cite{terzidou2023automated}.

Several automated systems and methodologies have been developed for this domain. For example, \cite{Itani2024automated} presented an automated process for anonymizing parole hearing transcripts in California, which reliably removes and pseudonymizes data while preserving structure. The ANOPPI tool \cite{Oksanen2022anoppi}, mentioned earlier, was specifically developed for Finnish legal texts. For German legal court decisions, an ML approach utilizing deep neural networks has been proposed for the automatic identification of sensitive text elements \cite{glaser2021anonymization}. Efforts also extend to scanned documents, with systems designed for the automatic anonymization of images of Swiss Federal Supreme Court rulings \cite{niklaus2023automatic} and more broadly for law enforcement documents, where machine learning is used to minimize manual effort and the extent of redacted areas \cite{eberhardinger2025anonymization}. Furthermore, \cite{CabreraDiego2024psilence} introduced PSILENCE, a pseudonymization tool for international arbitration documents that employs NER and coreference resolution for consistent entity replacement.

Highlighting the methodological needs, \cite{Csanyi2021challenges} reviewed the specific challenges of legal document anonymization, emphasizing that Named Entity Recognition (NER) alone is often insufficient and advocating for ML-based methods to complement it.
The crucial issue of re-identification continues to be a research focus. The capabilities of Large Language Models (LLMs) to re-identify individuals in anonymized court decisions were assessed by \cite{Nyffenegger2024anonymity}, who found current risks to be low for well-anonymized Swiss court cases but acknowledged the potential future threat. Investigating deanonymization further, \cite{beltrame2024redactbuster} introduced RedactBuster, a model that leverages sentence context to perform Named Entity Recognition on redacted documents. Tested on the Text Anonymization Benchmark (TAB), RedactBuster demonstrated high accuracy and also proposed countermeasures to enhance the privacy of redacted information.

\subsection{Audio, Call Centers, and Spoken Conversations}
Anonymizing spoken language introduces the complexity of Automatic Speech Recognition (ASR) errors and disfluencies. \cite{Baril2022named} presented a pipeline for anonymizing French audio data by using a forced aligner and an NER model to identify named entities in transcripts, then replacing corresponding audio segments with silence. For real-time applications, \cite{Gouvea2023trustera} introduced Trustera, a system that redacts PII in live spoken conversations in call centers by integrating ASR, NLU, and a live audio redactor. \cite{Kaplan2020may}, as noted before, focused on applying NER to call center transcripts to aid privacy law compliance.

\cite{davidson-etal-2021-improved} explored the application of state-of-the-art NER algorithms to ASR-generated call center transcripts, creating models with low latency that can be readily integrated into existing pipelines. In the context of call center transcripts, \cite{zhang2024data} presented a process for anonymizing training data and a framework for assessing its effectiveness, comparing models fine-tuned on anonymized data with commercially available LLM APIs. \cite{indonesian2025deidentification} developed and evaluated an efficient speech de-identification system for Indonesian speech in low-resource transcripts, using speech recognition, information extraction, and masking. 

\subsection{Educational Data}
Online learning environments generate vast amounts of student data, including discussion forum posts and essays, which require anonymization for research. \cite{Bosch2020hello} developed a method for discovering and redacting private information in online course forum texts. \cite{Farrow2023names} specifically tackled the challenge of pseudonymizing names, nicknames, and spelling errors in student discussions. \cite{Holmes2023deidentifying} explored a hybrid approach combining rule-based methods and a fine-tuned RoBERTa \cite{liu2019roberta} model to de-identify student writing, aiming to balance privacy and data utility for educational researchers.

Beyond direct redaction of PII from text, the field is also exploring broader privacy-enhancing technologies for educational data, which often incorporates or is derived from sensitive textual student inputs. \cite{chicaiza2020application} highlighted the relatively low adoption of anonymization techniques in Learning Analytics despite their crucial role, while also demonstrating how such techniques can be effectively integrated. Recognizing that traditional anonymization methods can be insufficient for the complexities of educational data, recent work has focused on more robust protection. For instance, \cite{liu2025advancing} proposed a Differential Privacy framework specifically designed for LA, offering practical guidance for its implementation and validating its effectiveness in safeguarding data privacy against potential attacks, while also exploring the trade-offs between privacy and utility. Another established technique, K-anonymity, was applied by \cite{triayudi2023educational} to student data in Educational Data Mining patterns, showing its potential for protecting student privacy while noting the decrease in correlation coefficients that must be balanced with information loss. Furthermore, the generation of privacy-preserving synthetic educational data, as explored by \cite{vie2022privacy}, offers an alternative to anonymizing original datasets. This approach particularly addresses the risks of re-identification from naively pseudonymized data and includes an evaluation framework for comparing synthetic data generators and techniques to guarantee privacy.

\subsection{Financial Reports}
The financial sector, characterized by highly confidential data, presents a critical domain for textual anonymization. Efforts here aim to enable the use of financial documents for tasks like text classification and entity detection without compromising sensitive information. \cite{biesner2022anonymization} specifically addressed the anonymization of German financial and legal documents by developing and evaluating methods based on neural network language models, including recurrent neural nets and transformer architectures. Their work also resulted in a web-based application for anonymizing such documents, demonstrating a practical approach to handling sensitive data. Building on the need for efficient solutions, \cite{Deusser2025resource} proposed a resource-efficient anonymization pipeline using knowledge distillation. This method transfers knowledge from large language models (LLMs) to smaller, more deployable encoder-only models, combined with named entity recognition and regular expressions, aiming to make anonymization scalable and accessible even without extensive computational resources or manually labeled data.

Beyond directly redacting information from documents, research also explores integrating privacy-preserving techniques into the analytical models themselves. For instance, \cite{basu-etal-2021-privacy} focused on privacy-enabled financial text classification. They proposed integrating Differential Privacy and Federated Learning with transformer-based models (BERT \cite{devlin-etal-2019-bert} and RoBERTa \cite{liu2019roberta}) to train on sensitive financial data while preserving privacy, highlighting the crucial trade-offs between privacy and utility in this domain. These works underscore the dual challenge in the financial sector: robustly anonymizing textual data and ensuring that subsequent analytical processes maintain privacy.

\section{Advanced Methodologies and Privacy-Preserving Techniques}

Beyond foundational NER and direct LLM applications, a significant body of research focuses on developing more sophisticated anonymization methodologies. These often incorporate explicit privacy models, risk assessment, advanced machine learning, and techniques aimed at better preserving data utility while enhancing privacy.

One line of research aims to integrate explicit privacy risk measures into the anonymization process. \cite{Papadopoulou2022neural} presented a three-step approach involving a privacy-enhanced entity recognizer, privacy-risk assessment measures (based on BERT \cite{devlin-etal-2019-bert}, web search, or classifiers), and linear optimization to mask entities while minimizing semantic loss under a risk threshold. \cite{ManzanaresSalor2025enhancingKBS} proposed enhancing NER-based anonymization by using explainability techniques with a neural language model to iteratively detect and mask terms posing the greatest re-identification risk until a user-defined k-anonymity level is reached. \cite{Lison2021anonymisation} argued for moving beyond simple sequence labeling by incorporating such explicit disclosure risk measures. \cite{Pilan2024truthful} presented a two-stage sanitization strategy using instruction-tuned LLMs to generate truth-preserving replacements for sensitive spans, then simulating inference attacks to select the most informative yet risk-resistant candidate.

Differential Privacy (DP) offers formal privacy guarantees and has been explored for text rewriting. \cite{Igamberdiev2022dprewrite} introduced DP-Rewrite, an open-source framework for reproducible DP text rewriting experiments. \cite{Igamberdiev2023dpbart} later proposed DP-BART, a system for Local Differential Privacy (LDP) text rewriting that significantly improved utility over prior methods. \cite{Meisenbacher2024dpmlm} introduced DP-MLM, using a masked language model for DP text rewriting, claiming better context preservation and utility at lower privacy budgets. \cite{Weggenmann2022dpvae} presented DP-VAE, using a differentially private Variational Autoencoder to generate human-readable, privatized versions of online reviews.

Other novel approaches include bootstrapping anonymization models. \cite{Papadopoulou2022bootstrapping} proposed using distant supervision from a knowledge graph to automatically annotate texts for k-anonymity, then fine-tuning a transformer model on this data. \cite{Hassan2023utility} explored word embedding-based anonymization, where risky terms are identified based on semantic similarity to a target profile and replaced with more general terms to preserve utility.

These advanced methodologies represent a shift towards more principled and robust anonymization, often striving for quantifiable privacy guarantees and a more nuanced balance between privacy protection and the utility of the anonymized data.

\section{Authorship Anonymization}

A distinct subfield within text anonymization is authorship anonymization, also known as authorship obfuscation. The goal here is not primarily to remove PII, but to modify the writing style of a text to prevent an author from being identified through linguistic patterns. This is crucial in contexts where an author wishes to remain anonymous, even if the content itself is not sensitive.

Several recent works have tackled this challenge using diverse techniques. \cite{Panov2022mucaat} proposed MuCAAT, a multilingual contextualized authorship anonymization method for social media texts, designed to alter stylistic fingerprints while preserving the message. Reinforcement learning has emerged as a promising approach: \cite{Loiseau2025tarot} introduced TAROT, an unsupervised method using policy optimization to regenerate texts, fooling an author classifier while maintaining task utility.

Constrained decoding with smaller language models is another avenue. \cite{Fisher2024jamdec} presented JAMDEC, an inference-time algorithm that uses constrained decoding (e.g., with GPT-2 XL) to produce stylistic variations, outperforming prior methods with similar-sized models and even competing with much larger models in obfuscation effectiveness. 

For languages other than English, research is also progressing. \cite{Franco2024evaluation} evaluated GAN-based and sequence-to-sequence models for authorship obfuscation of Portuguese texts, finding the GAN approach offered a better trade-off between classifier fooling and content preservation.

Authorship anonymization addresses a different facet of privacy than PII redaction, focusing on the implicit "signature" an author leaves in their writing. The development of these techniques is vital for protecting authors in sensitive situations and for ensuring freedom of expression.

\section{Evaluation Frameworks, Metrics, and Benchmarks}

The ability to reliably evaluate the effectiveness of anonymization techniques is paramount. This involves assessing both the level of privacy protection achieved and the extent to which data utility is preserved. Recent research has focused on developing dedicated corpora, robust metrics, and realistic evaluation scenarios, including attacker models.

A significant contribution is The Text Anonymization Benchmark (TAB) \cite{Pilan2022tabCL}, an open-source corpus of English-language court cases with comprehensive annotations of personal information, going beyond traditional de-identification by marking all spans needing masking. TAB also proposes evaluation metrics tailored for privacy protection and utility preservation.

However, traditional recall-based metrics have limitations. \cite{ManzanaresSalor2024evaluatingDMKD, ManzanaresSalor2022automaticPSD} argued against relying solely on comparison to a single ground truth and proposed evaluating residual disclosure risk via an automated re-identification attack, formalized as a multi-class classification problem leveraging neural language models. This approach directly measures how anonymous a text truly is against an intelligent adversary. \cite{Mozes2021nointruder} also argued for better evaluation criteria, proposing TILD (Technical performance, Information Loss, De-identification by humans) to standardize assessment.

The impact of anonymization on downstream NLP tasks is another critical evaluation dimension. \cite{Yermilov2023privacy} investigated how different pseudonymization techniques affect text classification and summarization, exploring the trade-offs between data protection and utility.

Benchmarking different approaches is also crucial. \cite{Asimopoulos2024benchmarkingMOCAST} provided a comparative study of transformer-based models and LLMs against traditional architectures for text anonymization using the CoNLL-2003 dataset, highlighting relative strengths and weaknesses. \cite{Singh2025unmasking} curated a challenging dataset with semi-synthetic sentences containing diverse PII types to expose performance gaps in widely used PII masking models, calling for better evaluation metrics and model transparency.

The re-identification capabilities of LLMs themselves are now part of the evaluation landscape. \cite{Nyffenegger2024anonymity}, as mentioned, assessed LLMs' ability to re-identify individuals in anonymized court decisions, providing insights into current risk levels. \cite{ManzanaresSalor2025enhancingKBS} also leverage re-identification risk within their enhancement methodology, implicitly evaluating against potential k-anonymity violations.

These efforts in developing robust evaluation frameworks are essential for advancing the field, ensuring that new anonymization methods provide genuine privacy protection while remaining practical for real-world applications.

\section{Anonymization Toolkits and Systems}

Beyond theoretical advancements and evaluation, the development of practical, usable toolkits and systems is paramount for deploying anonymization solutions in real-world scenarios. Several papers present such systems, often embodying the techniques discussed in previous sections.

For legal documents, ANOPPI \cite{Oksanen2022anoppi} offers a semi-automatic anonymization tool for Finnish texts, available as both a web application with a UI for human review and a REST API. PSILENCE \cite{CabreraDiego2024psilence} is another tool focused on legal texts, specifically for pseudonymizing international arbitration documents in English, emphasizing consistent entity replacement. In the clinical domain, INCOGNITUS \cite{Ribeiro2023incognitus} provides a flexible platform for automated anonymization of clinical notes, incorporating multiple techniques and a performance-evaluation module. For spoken conversations, particularly in call centers, Trustera \cite{Gouvea2023trustera} is a system that redacts PII in real-time, preventing human agents from hearing sensitive details while capturing the PII for authorized uses.

More general-purpose tools and libraries have also been developed to support broader anonymization needs. Microsoft's Presidio \cite{MsPresidio} is a context-aware, pluggable, and customizable PII anonymization service designed for text and images, offering a robust framework for detecting and protecting sensitive information across various applications. PII-Codex \cite{rosado2023pii} is a Python library designed for PII detection, categorization, and severity assessment, integrating with detection software to help users understand and manage the PII present in texts. Textwash \cite{Kleinberg2022textwash} is another Python tool combining NER, pattern matching, and fuzzy string matching for language-agnostic text anonymization. For researchers working with differentially private text rewriting, DP-Rewrite \cite{Igamberdiev2022dprewrite} provides a modular and extensible open-source framework to facilitate reproducible experiments and ensure transparency.

\section{Discussion and Future Directions}

The landscape of text anonymization has evolved significantly, driven by advancements in NLP, particularly the rise of LLMs, and increasing regulatory and societal demands for data privacy. Several key trends and challenges emerge from the reviewed literature.

\subsection{The Dual Role of LLMs}
LLMs offer unprecedented capabilities for sophisticated anonymization, including nuanced understanding of context, fluent text generation for pseudonymization or generalization, zero-shot PII detection and transformation \cite{Staab2024largeICLRWorkshop, Staab2024largeICLR, Liu2023deidgpt_arxiv}, and the leveraging of their knowledge for more efficient models \cite{Deusser2025resource}. Simultaneously, they represent powerful adversaries capable of re-identifying individuals from subtly anonymized texts \cite{Patsakis2023man, Nyffenegger2024anonymity} and are one of the main reasons many strive to anonymize their data \cite{Deusser2025resource}. Future work must focus on developing LLM-based anonymization techniques that are robust against LLM-based attacks, potentially through adversarial training or by designing defenses that explicitly account for LLM inference capabilities and evaluation frameworks \cite{Yang2024rupta}.

\subsection{Balancing Privacy and Utility}
This remains a central challenge. Overly aggressive anonymization can render data useless for downstream tasks, while insufficient anonymization leaves individuals vulnerable. In 2021, \cite{Lison2021anonymisation} emphasized moving beyond simple redaction to incorporate explicit disclosure risk measures . However, current techniques increasingly allow for a configurable trade-off, for example, by optimizing against a risk threshold \cite{Papadopoulou2022neural} or a desired k-anonymity level \cite{ManzanaresSalor2025enhancingKBS}. Methods that preserve semantic structure and utility while ensuring privacy, such as through embedding-based generalization \cite{Hassan2023utility} or conditional randomization of private attributes \cite{Frikha2024incognitext}, are crucial. Evaluating this balance requires comprehensive metrics and frameworks that capture both aspects effectively \cite{Pilan2022tabCL, Yermilov2023privacy}.

\subsection{Beyond Explicit PII}
Many traditional methods focus on redacting explicit identifiers (names, addresses). However, quasi-identifiers, contextual information, and even linguistic style can also lead to re-identification. New methods are increasingly addressing this by considering broader re-identification risks from residual data \cite{ManzanaresSalor2024evaluatingDMKD}, preventing attribute inference attacks \cite{Frikha2024incognitext}, and modifying stylometric features to obscure authorship and prevent author re-identification \cite{Panov2022mucaat, Fisher2024jamdec, Loiseau2025tarot}. This holistic view is essential for comprehensive privacy.

\subsection{Formal Privacy Guarantees vs. Practical Anonymization}
Techniques like Differential Privacy (DP), explored for text rewriting \cite{Igamberdiev2023dpbart, Meisenbacher2024dpmlm, Igamberdiev2022dprewrite} and privacy-preserving data generation or analysis \cite{Weggenmann2022dpvae, liu2025advancing, basu-etal-2021-privacy}, offer formal, provable privacy guarantees. However, applying them effectively to complex, unstructured text while maintaining high utility can be challenging, often involving intricate model design and careful parameter tuning. Bridging the gap between theoretically sound privacy models and practical, high-utility text anonymization that is accessible to a wider range of practitioners remains an active area of research.

\subsection{Domain Adaptation and Robustness}
Anonymization models often require domain-specific tuning due to variations in PII types, language use, and regulatory contexts. Developing techniques that generalize well across diverse text types and languages with minimal labeled data, potentially through innovative data augmentation strategies \cite{Kim2024privacysafe} or knowledge distillation from larger models \cite{Deusser2025resource}, is important for wider applicability across varied domains like healthcare \cite{Kocaman2023beyond, Kovacevic2024review}, legal \cite{terzidou2023automated}, and finance \cite{biesner2022anonymization}. The robustness of anonymization against noisy inputs (e.g., ASR errors in call center transcripts \cite{Kaplan2020may, davidson-etal-2021-improved}, or informal language in user-generated text like educational forums \cite{Bosch2020hello}) also needs continuous improvement.

\subsection{Evaluation and Benchmarking}
The development of standardized benchmarks like TAB \cite{Pilan2022tabCL} and attacker-centric evaluation methodologies, such as simulating re-identification attacks \cite{ManzanaresSalor2024evaluatingDMKD, ManzanaresSalor2022automaticPSD, beltrame2024redactbuster} or using challenging datasets to expose model weaknesses \cite{Singh2025unmasking}, is vital. These approaches help move evaluation beyond simple recall against a single ground truth \cite{Mozes2021nointruder} towards a more realistic assessment of privacy risks. Future efforts should expand these resources to more languages and domains, and incorporate more sophisticated and adaptive attack models, including those leveraging LLMs. Transparency in model capabilities, limitations, and evaluation processes is also key for building trust and fostering responsible innovation \cite{Singh2025unmasking}.

\subsection{Resource Efficiency and Accessibility}
While powerful models achieve strong results, their computational cost can be prohibitive for many applications. Research into knowledge distillation \cite{Deusser2025resource} and the development of efficient, specialized smaller models \cite{gunay2024llmsintheloop2expertsmall} is important for deploying anonymization on edge devices or in resource-constrained environments. A growing ecosystem of open-source tools and platforms \cite{Kleinberg2022textwash, Igamberdiev2022dprewrite, Oksanen2022anoppi, CabreraDiego2024psilence, MsPresidio, rosado2023pii} plays a crucial role in democratizing access to anonymization technologies, fostering reproducible research, and enabling practical adoption.

\subsection{Open Research Questions}
Despite these advancements, several critical challenges and open research questions persist, demanding further investigation:
\begin{itemize}
    \item \textit{Multilingual and Low-Resource Anonymization:} While many tools and techniques focus on high-resource languages like English, robust anonymization for a wider array of languages, especially low-resource ones, remains a significant hurdle. The lack of standardized multilingual benchmarks further complicates comparative evaluation and progress.
    \item \textit{Tension between Formal Privacy and Textual Quality:} Formal methods like Differential Privacy offer strong guarantees but can sometimes lead to text that is unnatural, lacks fluency, or has significantly diminished utility. Striking a better balance between provable privacy and the preservation of semantic integrity and readability is crucial, particularly for generative anonymization techniques.
    \item \textit{Dynamic and Adaptive Anonymization:} Current anonymization is often static. Developing systems that can dynamically adapt the level and type of anonymization based on context, data sensitivity, user permissions, or evolving threat models is an open area.
    \item \textit{Explainability and Trustworthiness of Anonymization Models:} Especially with complex models like LLMs, understanding \textit{why} certain information was (or was not) anonymized is important for trust and debugging. Methods for explaining anonymization decisions are needed.
    \item \textit{Anonymization of Emerging Data Modalities:} While this survey focuses on text, the principles extend. However, specific challenges arise with multimodal data (text + image/audio), code, and other structured or semi-structured textual forms.
\end{itemize}
Addressing these multifaceted challenges will require continued interdisciplinary collaboration, innovative algorithmic development, and a commitment to robust, context-aware evaluation to ensure that the benefits of natural language processing and text data can be realized responsibly while safeguarding privacy and adhering to existing and future regulations.

\section{Conclusion}

In this survey, we detailed the landscape of textual data anonymization. We have traced the evolution from foundational NER-driven techniques to the transformative impact of Large Language Models. Our review spanned domain-specific challenges and solutions in areas such as healthcare, law, and finance, alongside explorations of advanced privacy-preserving methodologies like Differential Privacy and the specialized domain of authorship anonymization. Furthermore, we underscored the indispensable role of robust evaluation frameworks, metrics, benchmarks, and the development of practical toolkits for real-world applicability.

Several key themes have emerged: the profound influence of LLMs as both powerful anonymizers and sophisticated re-identification threats; the enduring challenge of achieving a nuanced balance between robust privacy protection and data utility; the expanding scope of anonymization to address implicit identifiers and stylometric features; and the critical need for rigorous, attacker-aware evaluation. The development of resource-efficient and accessible tools also remains paramount for broader adoption.

As the amount of textual data is only increasing \cite{rivery2025bigdata}, the imperative to protect individual privacy while still enabling beneficial data use will only intensify. Addressing this will likely steer future efforts towards several key technical advancements: developing anonymization techniques specifically hardened against sophisticated LLM-driven re-identification attacks; refining the privacy-utility trade-off with more dynamic, context-aware, and even personalized mechanisms; and creating more sophisticated methods to obscure not just explicit PII but also subtle quasi-identifiers and stylometric fingerprints that can compromise anonymity. Furthermore, enhancing the practical applicability of formal privacy models like Differential Privacy to unstructured text, improving the robustness of anonymization against noisy and diverse data sources, and fostering the development of more resource-efficient models for wider deployment are all critical avenues. Progress in these domains, alongside the continuous evolution of comprehensive evaluation benchmarks and robust, adaptive attacker models, will be paramount for the advancement of text anonymization into an even more reliable tool, capable of meeting the privacy challenges of the future.

\section*{Acknowledgments}

For this survey, Google Gemini 2.5 Pro \cite{geminiteam2025geminifamilyhighlycapable} and GPT-4o \cite{openai2024gpt4ocard} were used to refine text portions in all sections. Additionally, we leveraged these models to screen, filter, and summarize potential papers for inclusion in this survey.

\bibliographystyle{unsrtnat}
\bibliography{bib}

\end{document}